\documentclass{article}
\usepackage[preprint]{colm2026_conference}

\usepackage{microtype}
\usepackage{hyperref}
\usepackage{url}
\usepackage{booktabs}
\usepackage{multirow}
\usepackage{graphicx}
\usepackage[inline]{enumitem}

\definecolor{darkblue}{rgb}{0, 0, 0.5}
\hypersetup{colorlinks=true, citecolor=darkblue, linkcolor=darkblue, urlcolor=darkblue}

\usepackage{lineno}
\title{Stories of Your Life as Others: A Round-Trip Evaluation of LLM-Generated Life Stories Conditioned on Rich Psychometric Profiles}

\author{
Ben Wigler$^{\heartsuit}$ \quad Maria Tsfasman$^{\heartsuit}$ \quad Tiffany Matej Hrkalovic$^{\spadesuit}$ \\
$^{\heartsuit}$LoveMind AI 
$^{\spadesuit}$Jheronimus Academy of Data Science}

\begin{document}

\ifcolmsubmission
\linenumbers
\fi

\maketitle

\begin{abstract}
Personality traits are richly encoded in natural language, and large language models (LLMs) trained on human text can simulate personality when conditioned on persona descriptions. However, existing evaluations rely predominantly on questionnaire self-report by the conditioned model, are typically limited in architectural diversity, and rarely use real human psychometric data. Without addressing these limitations, it remains unclear whether personality conditioning produces psychometrically informative representations of individual differences or merely superficial alignment with trait descriptors. To test how robustly LLMs can encode personality into extended text, we introduce a round-trip evaluation paradigm: we condition LLMs on real psychometric profiles from 290 participants to generate first-person life story narratives, and then task independent LLMs to recover personality scores from those narratives alone. We show that personality scores can be recovered from the generated narratives at levels approaching human test-retest reliability (mean \textit{r} = 0.750, 85\% of the human ceiling), and that recovery is robust across 10 LLM narrative generators and 3 LLM personality scorers spanning 6 providers. Decomposing systematic biases reveals that scoring models achieve their accuracy while counteracting alignment-induced defaults. Content analysis of the generated narratives shows that personality conditioning produces behaviourally differentiated text: nine of ten coded features correlate significantly with the same features in participants' real conversations, and personality-driven emotional reactivity patterns in narratives replicate in real conversational data. These findings provide strong evidence that the personality-language relationship captured during pretraining supports robust round-trip encoding and decoding of individual differences — including characteristic emotional variability patterns that replicate in real human behaviour.
\end{abstract} 

\section{Introduction}

Personality traits are among the most robust predictors of human behaviour, influencing outcomes across domains including interpersonal relationships, health, and occupational performance \citep{Roberts2007, Ozer2006}. The \textit{lexical hypothesis}, which holds that personality-relevant information is encoded in natural language, has motivated decades of research linking linguistic behaviour to personality dimensions \citep{SaucerGoldberg1996, BoydPennebaker2017}. This relationship has been observed in informal contexts such as social media \citep{FastFunder2008} as well as in structured settings such as job interviews \citep{Dai2022} and self-narratives \citep{Oltmanns2025}. The emergence of large language models (LLMs) trained on vast corpora of human-generated text has created new possibilities for studying the relationship between personality and language. Several lines of work have explored this from different angles: measuring synthetic personality traits exhibited by LLMs under psychometric testing \citep{SerapioGarcia2025, Sorokovikova2024}, probing where personality representations are stored within model parameters \citep{Ju2025}, and conditioning LLMs on persona descriptions to simulate specific individuals \citep{Argyle2023, Park2024}. Recent evidence suggests that narratively immersive persona prompting produces a more faithful personality simulation \citep{Bai2025, Kang2025}, although validation across diverse systems and with real human data remains limited.

Existing work on personality and LLMs exhibits several limitations. First, most studies evaluate persona conditioning through questionnaire self-report, which \citet{Han2025personality} showed to be dissociated from behavioural output; whether this reflects a fundamental limitation or insufficient conditioning depth remains open (\S2.2). Second, grounding in real human psychometric data remains limited: many studies use synthetic profiles \citep{Kwok2024, Bai2025} or thin prompting \citep{Kwon2026llmstrulyembodyhuman}, and most employ Big Five rather than HEXACO, leaving Honesty-Humility (trait central to prosocial behaviour \citep{AshtonLee2014}) unexamined. Constructs beyond personality inventories (trust, social anxiety, dark traits) also shape conversation \citep{Hrkalovic2025} but remain untested in LLM conditioning. Third, cross-architecture validation is rare, leaving open whether findings are architecture-specific.

The present study addresses these gaps by introducing a round-trip evaluation paradigm. Rather than asking whether an LLM can report personality on a questionnaire, we ask whether personality information encoded in a detailed immersive personality prompt can survive transformation into an extended first-person narrative and be recovered by an independent model. To our knowledge, this is the first demonstration of personality signal surviving a multi-stage transformation pipeline grounded entirely in real human psychometric data, evaluated against a held-out human ceiling. We ground this evaluation in the PARSEL dataset \citep{Hrkalovic2025}, which provides HEXACO-60 profiles \citep{AshtonLee2009} and 9 additional validated scales for 290 participants, along with recorded conversations that allow us to bridge synthetic and real behavioural data. We address three research questions:
\vspace{-5pt}
\begin{description}
\item[RQ1] \textbf{Round-trip personality recovery:} To what extent can LLMs encode and recover personality and psychometric scales beyond personality through narrative?
\vspace{-3pt}
\item[RQ2] \textbf{Signal validation:} Is the recovered signal personality-relevant and not an artifact?
\vspace{-3pt}
\item[RQ3] \textbf{LLM and human behaviour comparison:} How well does the LLM narrative behaviour reflect real human conversational behaviour?
\end{description}
\vspace{-5pt}
By answering these questions, this paper provides the following contributions:
\begin{enumerate*}[label=\textbf{(\arabic*)}]
  \item We introduce a round-trip evaluation paradigm that encodes real psychometric profiles into extended life story narratives and recovers personality from them using an independent model, achieving recovery close to human test-retest reliability.
  \item We validate cross-architecture generality across 10 generators and 3 scorers spanning 6 providers, strongly indicating that personality recovery is a general property of richly conditioned LLMs.
  \item We provide evidence that the recovered signal reflects personality content rather than biographical leakage or questionnaire memorisation, through masked matching experiments with independent models.
  \item We decompose systematic biases in the scoring pipeline, characterising how alignment-induced personality defaults interact with persona conditioning.
  \item We demonstrate that personality conditioning produces behaviourally differentiated narratives whose content features correlate with real human conversational behaviour, including a novel emotional reactivity pattern that replicates across synthetic and real data.
\end{enumerate*}

Taken together, these contributions establish that the personality-language relationship captured during pretraining is sufficient to support robust round-trip encoding and decoding of individual differences, bridging the LLM persona conditioning literature \citep{Bai2025, Kang2025} with psychometric assessment methodology \citep{AshtonLee2009, Henry2022} in an evaluation framework grounded in real human data.

\section{Related Work}

\textbf{Personality prediction from text.} Inferring personality from text has a long history in computational linguistics and psychology. The lexical hypothesis posits that personality-relevant distinctions are encoded in natural language \citep{SaucerGoldberg1996}, and computational approaches have leveraged this to predict personality from social media posts, essays, and spoken language \citep{BoydPennebaker2017, FastFunder2008}. Previous work has recovered personality from real human text at modest but meaningful levels: \citet{Dai2022} fine-tuned InterviewBERT on over 58,000 interview responses ($r = 0.37$ with self-reported HEXACO), \citet{Oltmanns2025} fine-tuned RoBERTa on real-life McAdams-style narratives ($r \sim 0.40$, $N = 1{,}409$), \citet{Wright2026} used a zero-shot LLM ensemble on stream-of-thought writing ($r = 0.415$), and \citet{Speer2026} compared multiple NLP methods on open-ended responses ($N = 904$), with a fine-tuned LLM achieving $r = 0.53$. These approaches recover personality from real human text. Our pipeline instead generates and then recovers personality from synthetic narratives conditioned on psychometric profiles, meaning recovery rates reflect encoding fidelity rather than natural expressive variability and therefore are not directly comparable to the previously reported rates.

\textbf{LLM persona conditioning.} A growing body of work investigates how LLMs can be conditioned on personality profiles to produce role-played behaviour \citep{Bai2025, SerapioGarcia2025, Shi2025, Zhou2025}. Despite progress, three gaps persist. First, most persona studies use brief trait descriptors or single-sentence prompts, limiting conditioning depth. \citet{Kang2025} showed that extended McAdams-style backstories (~2,500 words) produce substantially deeper persona binding than short descriptions, and \citet{Venkit2026} call for richer conditioning beyond single personality inventories. Without sufficient conditioning depth, it remains unclear whether shallow persona fidelity reflects a model limitation or a prompt limitation. Second, persona conditioning studies have predominantly used synthetic or researcher-constructed profiles \citep{Kwok2024} rather than validated instruments administered to real participants. This makes it difficult to assess whether models can represent actual individual differences. Third, \citet{Han2025personality} demonstrated that persona conditioning shifts LLM self-reports on questionnaires but does not proportionally shift task behaviour, raising a ``personality illusion'' concern. However, their conditioning used single-sentence descriptions and evaluated consistency through economic tasks where human personality-task correlations are typically small ($r \approx 0.10$-$0.20$); richer prompting substantially improves persona fidelity \citep{Bai2025, Kang2025}, suggesting the illusion may reflect insufficient conditioning rather than a fundamental limitation. Our work addresses these gaps by conditioning on real psychometric profiles across multiple validated scales and evaluating recovery from extended narrative as well as correspondence with real conversational behaviour.

\textbf{Personality expression across communicative contexts.} Whole Trait Theory \citep{Fleeson2001, Fleeson2015} conceptualises traits as density distributions of states rather than fixed points, predicting that the same individual expresses personality differently depending on the behavioral context. This is supported by evidence that other-ratings and self-ratings show partial agreement varying by trait observability \citep{Vazire2010, Connelly2010, Funder1995}. \citet{Han2025context} demonstrated that this context-sensitivity extends to LLM-simulated personas, which exhibit systematically different linguistic signatures across conversational contexts even under identical conditioning. It remains unexplored, however, whether personality signal encoded in structured LLM-generated narratives corresponds to how real individuals with the same profiles express personality in spontaneous conversation.

\textbf{Life story interviews and personality.} The McAdams Life Story Interview (LSI) is a structured protocol that elicits personality through narrative themes rather than direct trait self-description \citep{McAdams2001, McAdams2011}. Its sequence of life chapters, key scenes, and future scripts produces text rich in personality-relevant content across multiple domains \citep{Lodi-Smith2010}. \citet{Speer2026} formalise the link between prompt structure and recovery accuracy ($r = .70$), mediated by trait-relevant narrative unit density, and \citet{Oltmanns2025} adapted the protocol for personality prediction from real spoken narratives. These studies address whether personality can be inferred from human-authored text -- the recognition side of the problem. Our work addresses the complementary generation side: whether LLMs can produce LSI narratives that accurately encode a target personality profile.

\section{Method}

\subsection{Dataset}

We use personality data from the PARSEL dataset \citep{Hrkalovic2025}, a multimodal corpus collected to study social perception and partner selection in cooperative settings. PARSEL contains data from 297 participants who engaged in pair-wise conversations and collaborative tasks. Each participant completed the HEXACO-60 personality inventory \citep{AshtonLee2009}, yielding scores on six domains: Honesty-Humility (HH), Emotionality (E), Extraversion (EX), Agreeableness (A), Conscientiousness (C), and Openness to Experience (OP). In addition, participants completed instruments measuring trust (4 subscales: Propensity, Ability, Integrity, and Benevolence), the Perth Personality and Thinking Style inventory (PPTS; 4 subscales: Cognitive Responsiveness, Affective Responsiveness, Egocentricity, and Interpersonal Manipulation; \citealt{Medlin2018}), and the Social Interaction Anxiety Scale (SIAS; \citealt{Mattick1998}). In total, each participant's psychometric profile comprises 15 dimensional scores (6 HEXACO + 9 beyond-HEXACO). For the present study, we use $N = 290$ participants for whom complete psychometric profiles are available. Of these, 248 also have at least three conversation transcripts and are used for comparison between human and LLM behaviour.

\subsection{Pipeline overview\label{stages}}

\textbf{Stage 1: Immersive personality prompt generation} (Claude Opus 4.6, Anthropic). The prompt generator receives a participant's complete psychometric profile (all 60 HEXACO items, 6 domain means, and 9 beyond-HEXACO scores), along with biographical facts from conversation transcripts and a brief appearance description from webcam snapshots. From these, the LLM generator produces an immersive personality prompt of approximately 1,000 words written in second person (``You are someone who...''), translating numerical scores into natural language descriptions of each personality domain at the facet level, including concrete behavioural examples, interpersonal style, and characteristic emotional patterns. Each prompt covers all 15 constructs. See Appendix~\ref{app:prompt_gen} for the prompt generator comparison and model selection rationale.

\textbf{Stage 2: LSI narrative generation.} The LSI narrative generator receives the immersive personality prompt and generates a 24-turn McAdams Life Story Interview, adapted from \citet{McAdams2001} (see Appendix~\ref{app:lsi_mods} for modifications). The model produces first-person responses role-playing the individual described in the prompt, at temperature 1.0. The primary generator is GPT-4.1 (OpenAI); for cross-architecture validation, we additionally generate full $N = 290$ narratives using Gemini 3 Flash (Google), Grok 4.1 Fast (xAI), and Mercury 2 (Inception) along with six additional models at a random subsample of $N = 154$ participants (Appendix~\ref{app:extended_gen}). All generators operate without extended reasoning.

\textbf{Stage 3: Blind personality scoring.} The personality scorer receives the generated LSI narrative with no access to the original profile or immersive personality prompt. It scores all 60 HEXACO items on a 5 point Likert scale in a single API call (see Appendix~\ref{app:B60} for scoring method selection), at temperature 0.3 with no chain-of-thought. Our scoring approach follows \citet{Wright2026}, adapted for efficiency. Items are aggregated into 6 domain means following the standard HEXACO-60 scoring key \citep{AshtonLee2009}. In a second API call, the same scorer rates 51 additional items covering 9 beyond-HEXACO constructs (Trust subscales, PPTS subscales, SIAS), aggregated into subscale means following each instrument's published scoring key. Three scorers are used: Sonnet 4.6 (Anthropic), GPT-5.4 (OpenAI), and Gemini 3 Flash (Google).

\subsection{Signal validation controls}

\textbf{Masked matching.} To test whether narratives can be matched to their source profiles based on personality content alone, we constructed a forced-choice matching experiment. Gemini 3 Flash stripped all biographical information from the immersive personality prompts (the prompts produced in Stage 1, see Section \ref{stages} for details), leaving only personality descriptions. An independent model (Claude Haiku 4.5) verified that no biographical details remained in the stripped profiles. For each participant, three 5-option lineups were created with the correct stripped profile alongside four randomly selected stripped profiles from other participants. Three independent matchers (Claude Haiku 4.5, Grok 4.1 Fast, Gemini 3 Flash) attempted to match each narrative to its masked profile.

\textbf{Questionnaire leakage.} We tested whether narratives reproduce HEXACO-60 questionnaire items verbatim by computing sentence-level \textit{Jaccard} similarity between all narrative sentences and all 60 HEXACO item stems (chosen over embedding-based metrics as a conservative detector of near-verbatim reproduction). Sentences exceeding Jaccard $> 0.7$ were flagged as directly paraphrasing questionnaire items.

\textbf{Bias decomposition.} To understand where distortions enter the pipeline, we decompose total scoring bias into three additive stages. \textit{Stage 1 (Prompt generation bias):} Distortion introduced when converting raw psychometric scores into immersive personality prompts, measured as the difference between scores recovered directly from prompts and ground truth. \textit{Stage 2 (Narrative + scoring bias):} Distortion introduced during narrative generation and blind scoring. We decompose Stage 2 further by evaluating unconditioned narratives-life stories generated with no personality conditioning. \textit{Stage 2a (Pipeline resting bias):} The scores from unconditioned narratives represent the personality the scorer attributes when no signal is present, characterised using unconditioned narratives and self-reports from three generators (details in Appendix~\ref{app:distortion}). \textit{Stage 2b (Conditioning signal):} The difference between total Stage 2 distortion and resting bias, representing the directional pull of conditioning against the scorer's defaults.

\subsection{Content feature extraction and behaviour analysis\label{scoring_method}}

To test whether personality-conditioned LSI narratives produce behavioural signatures that reflect how real humans with the same profiles behave in conversation, we extracted content features from both generated LSI narratives and real PARSEL conversation transcripts. Ten features were coded: agency, communion, emotional intensity, vulnerability, disclosure depth, humor, warmth, dominance, emotional valence, and emotional complexity, all rated on a 5-point Likert scale. For narratives, each of the 24 McAdams sections was coded independently, yielding section-level profiles that enable analysis of within-narrative variance (i.e., how much a feature fluctuates across the interview). Three independent annotators coded all texts: Gemini 3 Flash (Google), Claude Haiku 4.5 (Anthropic), and GPT-5.4-Mini (OpenAI). Scores were averaged across annotators to yield a single estimate per unit. For narrative section-level coding, the Gemini-Haiku pair achieved mean ICC(2,1) = 0.762, indicating good reliability according to \citet{Cicchetti1994} (three-way ICC comparable). For conversation coding (577 conversations), three-way ICC = 0.483, with the best pair (Gemini-Haiku) at 0.568. For participants appearing in multiple conversations, speaker-level scores were averaged to yield a single participant-level estimate.

\section{Results}

\subsection{Round-trip personality recovery (RQ1)}

\begin{figure}[h!]
\centering
\includegraphics[width=0.9\textwidth]{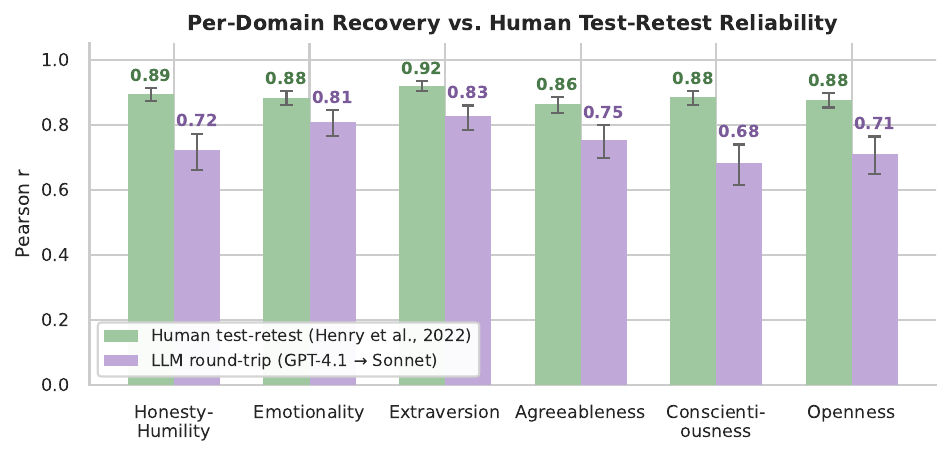}
\caption{Per-domain HEXACO recovery (teal; GPT-4.1 generator, Sonnet scorer, $N = 290$) vs.\ human test-retest reliability (rose; Henry et al., 2022, HEXACO-100, $N = 416$). Error bars: 95\% CI via Fisher $z$-transform.}
\label{fig:domain_comparison}
\end{figure}

\textbf{HEXACO recovery.} To quantify round-trip personality recovery, we compute Pearson $r$ between ground-truth HEXACO-60 domain means (participant self-report) and recovered domain means (blind scoring of generated narratives) across $N = 290$ participants. All correlations in \S4.1 are evaluated against a Bonferroni-corrected threshold of $\alpha = 0.0033$ (15 tests: 6 HEXACO + 9 beyond-HEXACO). The primary pipeline (GPT-4.1 generator, Sonnet scorer) yields a mean domain correlation of $r = 0.750$ (bootstrap 95\% CI [0.730, 0.768]). Per-domain recovery ranges from $r = 0.682$ (Conscientiousness) to $r = 0.825$ (Extraversion), representing 77-94\% of the human ceiling, all significant at $\alpha = 0.0033$ (Figure~\ref{fig:domain_comparison}). Honesty-Humility, Conscientiousness, and Openness show the largest gaps relative to human test-retest baseline, consistent with these domains' lower observability in LLM-generated LSI narratives. For context, human test-retest reliability for the HEXACO-100 is $r = 0.86$-$0.92$ across domains (median $r = 0.887$; \citealt{Henry2022}; $N = 416$, 13-day retest interval; verified by recomputing from published raw data, OSF: \url{https://osf.io/wz3du/}). The profile ceiling, obtained by scoring the immersive personality prompt directly without the narrative step, yields $r = 0.882$ (Sonnet) and $r = 0.878$ (Gemini 3 Flash), indicating that narrative generation reduces recovery by $\Delta r = 0.132$ while preserving the majority of recoverable signal.

\vspace{6pt}

\textbf{Beyond-HEXACO signal recovery.} All 9 beyond-HEXACO subscales (4 Trust, 4 PPTS, SIAS) are recovered at statistically significant levels (range: $r = 0.314$-$0.645$, all $p < 0.0033$; full results in Appendix~\ref{app:beyond_hexaco}). Interpersonal and affective constructs (Trust Benevolence $r = 0.645$, PPTS Affective Responsiveness $r = 0.594$) recover most strongly; cognitive and propensity measures recover more weakly (PPTS Cognitive Responsiveness $r = 0.314$, Trust Propensity $r = 0.420$).

\begin{figure}[h!]
\centering
\includegraphics[width=0.9\textwidth]{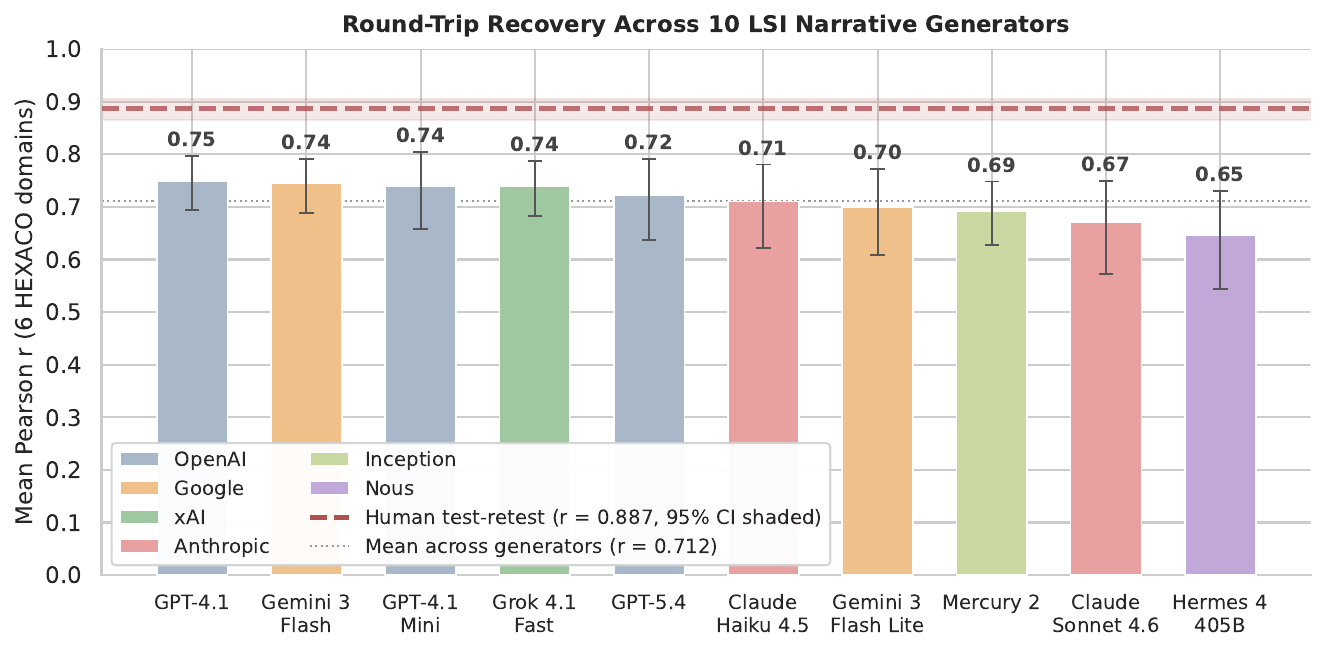}
\caption{Round-trip recovery across generators from 6 providers (including Mercury 2, a diffusion LM), scored by 3 independent scorers. Bars: mean $r$ across 6 HEXACO domains ($N = 290$ for primary generators, $N = 154$ for extended set), colour-coded by provider. Dashed line: human test-retest (95\% CI shaded). Error bars: 95\% CI. Full results in Appendix~\ref{app:extended_gen}.}
\label{fig:model_comparison}
\end{figure}

\textbf{Cross-architecture validation.} Three primary generators at $N = 290$ (GPT-4.1, Gemini 3 Flash, Grok 4.1 Fast) yield Sonnet-scored recovery of $r = 0.750$, $0.744$, and $0.740$; nine generator-scorer pairs yield correlations spanning within $r = 0.719$-$0.750$. Mercury 2 (Inception Labs), a diffusion language model tested at $N = 290$, achieves $r = 0.693$-$0.697$ (within 6 points of frontier autoregressive generators) suggesting personality-language associations arise from training data rather than autoregressive generation specifically. Extended testing at a subsample of $N = 154$ across six additional generators confirms generality (Appendix~\ref{app:extended_gen}). As Figure~\ref{fig:model_comparison} shows, personality recovery is robust across architectures and generation paradigms (see Appendix~\ref{app:sonnet_gen} for Sonnet's atypical generator performance).

\subsection{Signal validation (RQ2)}

\textbf{Masked matching.} Three independent matchers on 870 forced-choice trials (290 participants $\times$ 3 lineups, 5 options each; chance = 20\%): Claude Haiku 4.5 -- 79.4\% accuracy ($p < 0.001$); Grok 4.1 Fast -- 79.0\% accuracy ($p < 0.001$); Gemini 3 Flash -- 95.2\% accuracy ($p < 0.001$). Haiku and Grok  produce near-identical accuracy, differing by 0.4 percentage points; Gemini's higher accuracy likely reflects a same-model advantage (Gemini also performed the stripping), so we report Haiku/Grok as the primary result.

\textbf{Questionnaire leakage.} We searched all narratives (totalling \~{}2.3M words) for reproductions of HEXACO-60 item stems using sentence-level Jaccard similarity. Zero sentences exceeded the Jaccard $> 0.7$ threshold. While a mean of 7.9 self-descriptions per narrative (SD = 2.3) overlap thematically with item content, no exact or near-exact item reproductions were detected. The personality signal is expressed through embodied narrative content, not through parroted questionnaire items.

\textbf{Bias decomposition.} Applying the decomposition defined in \S3.3: \textbf{Stage~1} introduces positive bias on HH (+0.499) and C (+0.200). \textbf{Stage~2a} reveals that three unconditioned frontier models share a characteristic resting personality: HH$\approx$5.0, E$\approx$1.0-1.3, C$\approx$5.0, consistent with the ``helpful, harmless, honest'' alignment signature (full profiles in Appendix~\ref{app:distortion}). Narrative generation partially humanises this default (e.g., E rises 1.2-1.7 points), but models remain far from human population means. \textbf{Stage~2b} shows that conditioning operates \textit{against} these defaults: for HH in the GPT-4.1$\rightarrow$Sonnet pipeline, resting bias is +1.780 while conditioning corrects by $-$1.409. Across all pairs, 53\% of Stage~2 distortion is resting bias and 47\% is conditioning actively pulling toward ground truth. Two architecturally different scorers produce the same bias pattern ($r = 0.994$), confirming that distortion resides in shared training signals rather than scorer idiosyncrasies.

\subsection{Bridging synthetic and real behaviour (RQ3)}

To test whether the personality signal encoded in LSI narratives connects to how the same individuals behave in real conversation, we compared content features between generated LSI narratives and real PARSEL conversation transcripts for the $N = 248$ participants who had both narrative coding and at least three conversations. For each of the 10 features, we computed Pearson $r$ between narrative-derived and conversation-derived scores (Bonferroni-corrected $\alpha = 0.005$, 10 tests). Nine of ten features show significant positive human-LLM behaviour correlations (range $r = 0.131$-$0.268$; top: vulnerability $r = 0.268$, agency $r = 0.245$, emotional valence $r = 0.218$; only humor non-significant); mean $|r| = 0.184$.

Beyond mean levels, personality predicts \textit{feature variance within the generated LSI narratives}: the standard deviation of emotional valence across the 24 interview sections correlates more strongly with Emotionality ($r = 0.303$, $p < 0.001$) than mean emotional valence does ($r = -0.231$), indicating that high-Emotionality profiles produce narratives that \textit{oscillate} more between positive and negative sections. This reactivity pattern replicates in real conversations: within-person valence variance across conversation partners also correlates with Emotionality ($r = 0.170$, $p = 0.007$), confirming that the LLM-generated pattern reflects a real human behavioural signature. Correlating LLM-coded narrative features with ground-truth HEXACO scores, Emotionality predicts vulnerability ($r = 0.744$) and emotional intensity ($r = 0.576$), Agreeableness predicts warmth ($r = 0.584$) and communion ($r = 0.526$), Honesty-Humility predicts meaning-making ($r = 0.499$), and Extraversion predicts agency ($r = 0.494$), confirming personality conditioning shapes narrative content in theoretically expected directions (full results in Appendix~\ref{app:convergent}). The feature-personality mapping and reactivity finding replicate across all four generators including Mercury 2 ($r = 0.249$; Appendix~\ref{app:convergent}). Unsupervised topic modelling (BERTopic) independently confirms that personality organises LLM-generated LSI narrative structure at the embedding level, with 37 significant topic-personality associations emerging without supervised feature extraction (Appendix~\ref{app:bertopic}).

\begin{figure}[h!]
\centering
\includegraphics[width=0.9\textwidth]{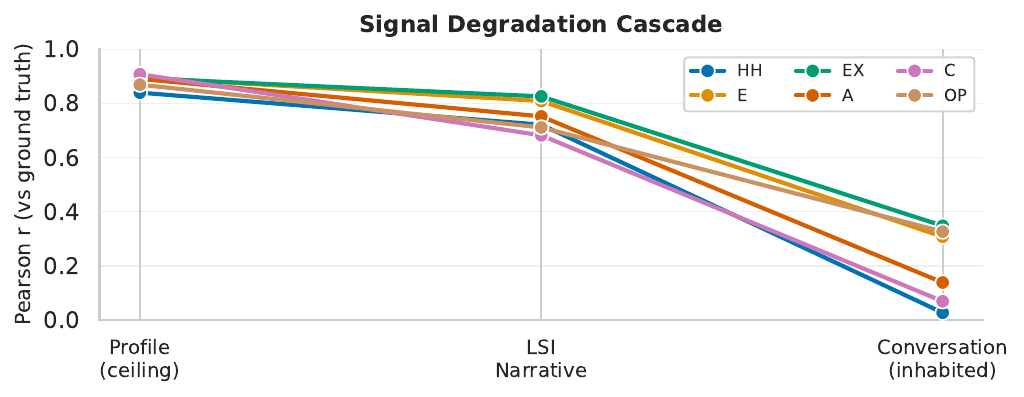}
\caption{Signal degradation across three stages: profile ceiling (Sonnet on prose profiles, $N = 290$; $r = 0.839$-$0.907$), LSI narrative recovery (GPT-4.1 generator, $N = 290$; $r = 0.682$-$0.825$), and real conversation (Spearman $\rho$, $N = 248$; $\rho = 0.046$-$0.418$). }
\label{fig:degradation_cascade}
\end{figure}

\section{Discussion}

\textbf{RQ1: Recovery and generalisation.} Our primary finding is that personality scores encoded into extended life story narratives can be recovered by independent LLMs at $r = 0.750$, representing 85\% of human test-retest reliability \citep{Henry2022}. Two features of this result warrant emphasis. First, the signal generalises across independently trained models: three generators from three providers yield Sonnet-scored recovery of $r = 0.750$, $0.744$, and $0.740$, and 8 of 9 frontier-class models tested exceed $r = 0.70$. This consistency suggests that personality-language associations are a general property of large-scale language modelling rather than model-specific artifacts, in line with \citet{Pellert2026}, who showed that LLM embedding spaces encode psychometric factor structure without human response data. Recovery by Mercury 2, a diffusion language model with a fundamentally different generation mechanism, further indicates that this property is rooted in training data rather than the autoregressive inductive bias shared by all other tested models. Second, recovery extends beyond HEXACO to all 9 beyond-personality subscales. Interpersonal and affective constructs (Trust Benevolence, PPTS Affective, Egocentricity) are recovered most strongly, while cognitive and propensity-based constructs recover more weakly -- a pattern that may reflect the LSI format's emphasis on interpersonal scenarios and emotional experiences, providing more opportunity for affective traits to be expressed.

\textbf{RQ2: Signal validity.} Independent LLMs matched narratives to personality-only source profiles (all biographical details removed) at ~79\% accuracy (chance: 20\%), confirming that the narratives encode personality signal beyond surface-level cues. The zero verbatim leakage result indicates that conditioned LLMs express personality through embodied content (scenes, emotional reactions, interpersonal dynamics, and moral reasoning) rather than paraphrased questionnaire items. The pipeline mitigates memorisation at every stage: immersive personality prompts avoid item-level language, the 24-turn LSI format requires sustained first-person narrative, and the scoring model has no access to the original profile. The bias decomposition provides further evidence: unconditioned frontier LLMs produce a shared resting profile (HH = 5.0, C = 5.0, E = 1.0) reflecting the alignment signature, and conditioned narratives partially overcome these defaults, suggesting that recovery accuracy depends not only on conditioning richness \citep{Bai2025} but also on the distance between target personality and scorer default.

\textbf{RQ3: Bridging synthetic and real behaviour.} Nine of ten content features extracted from personality-conditioned LSI narratives show significant positive correlations with the same features in participants' real conversations (mean $|r| = 0.184$, range $0.131$--$0.268$), providing convergent evidence that the narrative content reflects genuine behavioural dispositions. The moderate magnitude is expected: LSI narratives are ${\sim}$8,000-word structured life reviews eliciting reflective self-narration, while the conversations are 3-minute cooperative exchanges governed by social dynamics. Personality predicts not only mean feature levels but within-narrative variance: high-Emotionality profiles produce narratives that oscillate more in emotional valence across sections, and the same reactivity pattern appears in real conversations (within-person valence variance across partners). This is consistent with Whole Trait Theory's conceptualisation of traits as density distributions of states rather than fixed points \citep{Fleeson2001, Fleeson2015}: the pipeline captures not only trait central tendency but characteristic \textit{variability}. The same feature-personality mapping replicates across all three primary generators, confirming that behavioural differentiation is a property of the conditioning pipeline rather than any single model (Appendix~\ref{app:convergent}).

\textbf{Comparison to existing benchmarks.} Our zero-shot recovery ($r = 0.750$) exceeds prior benchmarks from natural text: $r = 0.37$ from supervised interview coding \citep{Dai2022}, $r \sim 0.40$ from real-life narratives \citep{Oltmanns2025}, $r = 0.42$ from zero-shot LLM ensembles \citep{Wright2026}, and $r = 0.53$ from fine-tuned LLMs \citep{Speer2026}. However, direct comparison is limited by a fundamental structural difference: our pipeline generates personality-expressive text, while prior work extracts personality from natural text where it is one of many signals.

\textbf{Limitations.} Several limitations should be noted. First, the round-trip design evaluates whether personality information can be preserved through narrative transformation, not whether it is present in natural human text to the same degree. Second, all content feature coding is performed by LLMs rather than trained human annotators, introducing a shared-method concern; future work would greatly benefit from ground-truth human annotation. Third, inter-annotator reliability for conversation coding (three-annotator ICC = 0.483) is lower than for narrative coding (ICC = 0.762), reflecting the difficulty of coding short conversational texts. Fourth, the PARSEL dataset represents a specific population (English-speaking adults in a cooperative task setting), and no human-authored LSI narratives from this population are available for direct comparison, precluding evaluation of how synthetic stories differ from human ones in quality or structure. Lastly, while cross-architecture validation demonstrates generality, all tested models share large-scale English web training data, and we cannot fully rule out shared distributional biases.

\vspace{-5pt}
\section{Conclusion}
\vspace{-5pt}

We set out to address three gaps in the literature on personality and large language models: the reliance on questionnaire self-report for evaluating persona conditioning, the lack of grounding in real human psychometric data, and the absence of cross-architecture validation. Using a three-stage round-trip pipeline (immersive personality prompt generation from real psychometric profiles, life story interview generation, and blind personality recovery), we demonstrate that personality information can survive multiple transformations at levels approaching human test-retest reliability ($r = 0.750$, 85\% of ceiling). This signal generalises across 10 generators and 3 scorers spanning 6 providers. It is robust beyond surface-level cues: after removing biographical identifiers, independent LLMs match narratives to source profiles at ~79\% accuracy based on personality content alone. And it is not disconnected from real behaviour: nine of ten content features in generated narratives correlate significantly with the same features in participants' actual conversations, and personality-driven emotional reactivity patterns in generated narratives replicate in real conversational data. Decomposing the pipeline's systematic biases reveals that scoring models achieve their accuracy while counteracting alignment-induced defaults (high Honesty-Humility, high Conscientiousness, low Emotionality), suggesting that the personality-language relationship encoded during pretraining is robust enough to overcome post-training distortions. These findings have implications for personality assessment, where the LSI format shows promise as a naturalistic alternative to questionnaire-based measurement; for persona conditioning in AI systems, where our results join growing evidence that richer immersive prompting substantially outperforms thin trait descriptions \citep{Bai2025, Kang2025}; and for understanding how personality is represented in language models, where our results, together with evidence that personality structure is encoded in embedding spaces \citep{Pellert2026} and that personality subnetworks are physically localisable in model parameters \citep{Ju2025}, suggest that pretraining on human text produces personality-language associations that are general, recoverable, and behaviourally relevant.

\newpage
\section*{Reproducibility Statement}

All psychometric data are from the PARSEL dataset, available upon request from the original authors \citep{Hrkalovic2025}. Model versions, prompts, scoring parameters, statistical analysis scripts, and results will be released upon acceptance.

\section*{Ethics Statement}

This study uses existing psychometric data collected under ethical approval ([anonymised]). No directly identifiable information appears in the generated narratives. We note that the capacity to recover personality from generated text raises privacy considerations that warrant careful attention in deployment contexts. Nevertheless, the demonstrated capacity to generate personality-consistent life narratives from psychometric data raises several ethical considerations.      

First, the ability to produce extended first-person text that reflects an individual's personality profile introduces impersonation risks. While the generated narratives do not reproduce real autobiographical content, they could in principle be used to create synthetic text that mimics the communicative style of a real person, given access to their psychometric profile. As personality questionnaires are widely administered in employment, clinical, and educational contexts, safeguards against misuse of such data for synthetic identity generation warrant careful consideration in downstream applications.
                                                                  
Second, the psychometric instruments used in this study predominantly originate from WEIRD (Western, Educated, Industrialised, Rich, and Democratic) populations. The HEXACO model, while cross-culturally validated, may not capture personality dimensions that are salient in non-Western cultural contexts. Additionally, the language models employed as both generators and scorers were primarily developed by US-based organisations and trained on English-dominant data, introducing a further layer of cultural bias in how personality traits are expressed and interpreted. As a result, the personality representations produced by our pipeline may inadvertently reinforce culturally specific behavioural norms as universal, while underrepresenting alternative expressions of the same underlying traits. Further research is needed to evaluate the robustness of personality-conditioned generation across languages, cultural backgrounds, and underrepresented populations.

Third, the experimental design involves generating and scoring narratives across multiple LLM architectures and conditions, resulting in a substantial computational footprint. We acknowledge the environmental cost of large-scale LLM inference and note that future work should consider efficiency-focused strategies, such as smaller open-weight models or targeted subset designs, to reduce the energy demands of personality-conditioned generation research.

\section*{AI Disclosure Statement}

As the generative pipeline is itself the subject of this research, large language models served as experimental instruments throughout: as prompt generators, narrative generators, and personality scorers. LLMs also served as content annotators and masked matchers, with inter-annotator reliability reported explicitly (see Section \ref{scoring_method}). The human authors were additionally aided in literature search, coding support, and manuscript preparation (e.g. grammar checking and paper shortening) by a generative agent operating within a customised coding harness augmented with long-term memory capabilities; this agent's system instructions can be made available upon request. Generated narratives were evaluated for personality signal recovery, not subjective quality. All methodological decisions, experimental design, and interpretive claims are by the human authors, who are solely responsible for the accuracy, academic integrity, and intellectual contributions of this paper.

\bibliography{references}
\bibliographystyle{colm2026_conference}

\appendix

\section{Content versus structural features}
\label{app:content}

Analysis of what within the narratives carries personality signal reveals a clear dissociation. Structural features (sentence length, word count, paragraph structure) show near-zero personality correlations. The only exception is type-token ratio, which correlates with Openness at $r = 0.355$, suggesting that open individuals receive broader vocabulary even in generated text. Sentence length coefficient of variation is 0.075 across participants in LSI narratives versus 1.066 in real conversations, confirming that the generator imposes uniform structural form regardless of personality.

Content features show strong correlations (all evaluated at Bonferroni-corrected $\alpha = 0.00024$, 204 tests: 34 features $\times$ 6 HEXACO domains): vulnerability with Emotionality at $r = 0.744^{***}$, emotional intensity with Emotionality at $r = 0.576^{***}$, warmth with Agreeableness at $r = 0.584^{***}$, and agency with Extraversion at $r = 0.494^{***}$. Of 204 feature--domain pairs tested, 55 survive Bonferroni correction ($N = 290$, GPT-4.1 narratives). An immersive personality prompt control, in which three independent LLM coders applied the same content rubric to both narratives and source prompts, shows that 54 out of 90 feature-domain pairs (60\%) exhibit stronger personality-content correlations in narratives than in prompts, with all three coders agreeing on narrative $>$ prompt for the majority of features.

\section{Sonnet as generator}
\label{app:sonnet_gen}

Sonnet 4.6 evaluated as a generator ($N = 154$) yields lower recovery across all scorers: Sonnet-to-GPT-5.4 $r = 0.670$, Sonnet-to-Gemini $r = 0.628$, mean $r = 0.650$. This consistent underperformance ($\Delta r \approx 0.10$ vs.\ GPT-4.1 and Gemini 3 Flash) is consistent with the scoring-generation dissociation: RLHF compression attenuates personality signal during generation while preserving---and possibly enhancing---scoring sensitivity. Sonnet is simultaneously the weakest frontier generator and the strongest scorer ($r = 0.750$).

\section{OLMo contamination control}
\label{app:olmo}

To rule out the possibility that personality recovery depends on PARSEL data contamination in frontier model training, we scored narratives using OLMo (Allen AI), an open-source model trained exclusively on the Dolma corpus, which does not contain PARSEL. OLMo recovery ($r = 0.611$) is lower than frontier scorers but substantially above chance, confirming that the personality signal in narratives is not an artifact of training data overlap.

\section{B10 vs B60 ablation details}
\label{app:ablation}

Domain-batched scoring (B10, $N = 50$): $r = 0.711$. All-at-once scoring (B60, $N = 50$): $r = 0.747$. Delta = 0.036. B10 with chain-of-thought: $r = 0.709$. B10 without CoT: $r = 0.711$. CoT provides no benefit and marginally reduces accuracy, possibly by introducing reasoning noise that overrides the scorer's implicit personality model.

\section{LSI modifications}
\label{app:lsi_mods}

Our LSI protocol is adapted from the McAdams Life Story Interview II (McAdams, 2007), preserving the core structure (life chapters, 8 key scenes, future script, challenges, personal ideology, life theme) while making four modifications for the LLM generation context. First, we replaced the religious/spiritual experience scene with an important adolescent scene, as frontier LLMs frequently refuse or produce formulaic responses to religious prompts. Second, we neutralized the valence specification for childhood memory prompts (McAdams specifies ``positive childhood memory'' and ``negative childhood memory''; ours requests ``earliest memory'' and ``important childhood scene'') to avoid eliciting content involving harm to minors. Third, we broadened the ``change in religious/political views'' prompt to ``personal growth,'' allowing a wider range of personality-relevant development narratives. Fourth, we added two reflexive closing questions---``How do you think others perceive you?'' and ``Do you notice any patterns or surprises in your story?''---to compensate for the absence of a live interviewer who would typically probe for self-reflection throughout the session.

\section{Scoring method selection}
\label{app:B60}

We evaluated scoring configurations on a pilot set of $N = 50$ participants. Scoring all 60 HEXACO items in a single API call (B60) yielded $r = 0.747$ versus $r = 0.711$ for domain-batched scoring (B10: 6 calls of ${\sim}$10 items each), a difference of $\Delta r = 0.036$. Adding chain-of-thought reasoning to B10 provided no improvement ($r = 0.709$). We attribute the B60 advantage to holistic context: the scorer can use cross-domain consistency when evaluating all items simultaneously. B60 is also 6$\times$ more cost-efficient. All results reported in this paper use the B60 method at temperature 0.3 with no chain-of-thought.

\section{Prompt generator comparison}
\label{app:prompt_gen}

We tested 5 alternative models (GPT-5.4, Hermes 4 405B, Hermes 4 70B, Aion 2.0, Gemma 3 27B) against Opus 4.6 on the 15 most challenging participant profiles (those most distant from RLHF defaults: low HH, low C, high E). After applying a standardised review step to all generators, GPT-5.4 ($r = 0.896$) and Hermes 405B ($r = 0.880$) approached Opus performance ($r = 0.910$), but only Opus consistently produced vivid, unhedged descriptions of socially undesirable traits across the full profile range. Gemma 3 27B---tested with and without RLHF (``abliterated'')---showed that removing safety training worsened rather than improved recovery for the target traits (RLHF: $r = 0.714$; abliterated: $r = 0.647$; HH specifically: 0.414 vs.\ 0.117), indicating that the dark-trait generation challenge is a capability problem, not a refusal problem. 

\section{Multi-stage distortion scan}
\label{app:distortion}

Three stages of personality distortion characterised across three generators (GPT-4.1, Gemini 3 Flash, Grok 4.1 Fast):

\paragraph{Unconditioned B60 self-report (raw HEXACO, temp=0):}

\begin{table}[h]
\begin{center}
\begin{tabular}{lcccccc}
\toprule
\textbf{Generator} & \textbf{HH} & \textbf{E} & \textbf{EX} & \textbf{A} & \textbf{C} & \textbf{OP} \\
\midrule
GPT-4.1 & 5.0 & 1.0 & 3.4 & 4.8 & 5.0 & 4.6 \\
Gemini Flash & 5.0 & 1.3 & 4.3 & 4.4 & 5.0 & 4.7 \\
Grok Fast & 5.0 & 1.0 & 4.8 & 4.5 & 4.9 & 4.6 \\
\textit{Human mean} & \textit{3.43} & \textit{3.29} & \textit{3.23} & \textit{3.29} & \textit{3.69} & \textit{3.69} \\
\bottomrule
\end{tabular}
\end{center}
\caption{Unconditioned B60 self-report scores (temperature = 0) compared to human population means.}
\label{tab:uncond_b60}
\end{table}

\paragraph{Unconditioned LSI (entity prompt, no personality, Sonnet-scored, temp=0):}

\begin{table}[h]
\begin{center}
\begin{tabular}{lcccccc}
\toprule
\textbf{Generator} & \textbf{HH} & \textbf{E} & \textbf{EX} & \textbf{A} & \textbf{C} & \textbf{OP} \\
\midrule
GPT-4.1 & 4.4 & 2.7 & 4.0 & 4.0 & 4.0 & 4.3 \\
Gemini Flash & 4.8 & 2.8 & 4.1 & 3.9 & 4.6 & 4.7 \\
Grok Fast & 4.6 & 2.2 & 4.7 & 3.6 & 4.1 & 4.6 \\
\bottomrule
\end{tabular}
\end{center}
\caption{Unconditioned LSI scores (entity prompt, no personality conditioning, Sonnet-scored, temperature = 0).}
\label{tab:uncond_lsi}
\end{table}

\paragraph{Unconditioned variance (SD across temp=1 runs):}

\begin{table}[h]
\begin{center}
\begin{tabular}{lcc}
\toprule
\textbf{Generator} & \textbf{B60 SD ($N$=10)} & \textbf{LSI SD ($N$=5)} \\
\midrule
GPT-4.1 & 0.115 & 0.115 \\
Gemini Flash & 0.093 & 0.159 \\
Grok Fast & 0.097 & 0.189 \\
\bottomrule
\end{tabular}
\end{center}
\caption{Mean SD across 6 HEXACO domains. B60 variance is the model's natural wobble on the questionnaire; LSI variance is the model's variability in generating and being scored on different life stories. Both are far smaller than conditioned variance (SD $\approx$ 0.55--0.75 across 290 PIDs), confirming that personality conditioning does real work.}
\label{tab:uncond_var}
\end{table}

\section{Extended generator comparison}
\label{app:extended_gen}

To characterise how personality recovery varies across model families and scales, we report results at two sample sizes. Three primary generators (GPT-4.1, Gemini 3 Flash, Grok 4.1 Fast) were run at $N = 290$; seven additional generators were run at $N = 154$. All scored by the canonical three-scorer team (Sonnet, GPT-5.4, Gemini; self-scoring excluded).

\begin{table}[h]
\begin{center}
\begin{tabular}{clccccc}
\toprule
\textbf{Rank} & \textbf{Generator} & \textbf{Provider} & \textbf{Sonnet} & \textbf{GPT-5.4} & \textbf{Gemini} & \textbf{Mean} \\
\midrule
1 & GPT-4.1$^\dagger$ & OpenAI & $\mathbf{0.750}$ & 0.738 & 0.728 & 0.739 \\
2 & Gemini 3 Flash$^\dagger$ & Google & 0.744 & 0.736 & --- & 0.740 \\
3 & Grok 4.1 Fast$^\dagger$ & xAI & 0.740 & 0.735 & 0.719 & 0.731 \\
\midrule
4 & GPT-4.1 Mini & OpenAI & 0.740 & 0.734 & 0.712 & 0.729 \\
5 & GPT-5.4 & OpenAI & 0.723 & --- & --- & 0.723 \\
6 & Mercury 2$^\dagger$ & Inception Labs & 0.693 & 0.697 & 0.654 & 0.681 \\
7 & Claude Haiku & Anthropic & 0.710 & 0.701 & 0.662 & 0.691 \\
8 & Gemini 3 Flash Lite & Google & 0.700 & 0.709 & 0.678 & 0.696 \\
9 & Claude Sonnet 4.6 & Anthropic & --- & 0.670 & 0.628 & 0.649 \\
10 & Hermes 4 405B & Nous & 0.647 & 0.639 & 0.629 & 0.638 \\
\bottomrule
\end{tabular}
\end{center}
\caption{Extended generator comparison. Mean $r$ across 6 HEXACO domains, self-scoring excluded. $^\dagger$Primary generators and Mercury 2 at $N = 290$; all others at $N = 154$. Mercury 2 is a diffusion language model; all other generators are autoregressive.}
\label{tab:extended_gen}
\end{table}

Three tiers emerge: (1) frontier autoregressive generators converge at $r = 0.72$--$0.75$ regardless of architecture; (2) smaller/efficiency models and Mercury 2 (diffusion) maintain $r = 0.68$--$0.71$; (3) open-source or heavily safety-trained generators fall to $r = 0.64$--$0.65$. GPT-4.1 Mini ($r = 0.740$ Sonnet-scored) nearly matches GPT-4.1 at a fraction of the inference cost. Mercury 2 ($r = 0.693$--$0.697$), a diffusion language model, achieves recovery within 5--6 points of frontier autoregressive models despite a fundamentally different generation mechanism, suggesting that personality-language associations emerge from training data rather than autoregressive inductive biases specifically. Sonnet 4.6 as generator ($r = 0.650$) exhibits the scoring-generation dissociation discussed in Appendix~\ref{app:sonnet_gen}.

\section{Beyond-HEXACO subscale recovery}
\label{app:beyond_hexaco}

In addition to the 6 HEXACO domains, participants completed 9 beyond-HEXACO subscales: 4 Trust subscales \citep{Mayer1999}, Social Interaction Anxiety (SIAS; \citealt{Mattick1998}), and 4 Psychopathic Personality Traits (PPTS; \citealt{Coid2009}) subscales. These scores were included in the immersive personality prompts and scored from generated narratives using published item sets.

\begin{table}[h]
\begin{center}
\begin{tabular}{lccc}
\toprule
\textbf{Subscale} & \textbf{$r$} & \textbf{$p$} & \textbf{$N$} \\
\midrule
Trust: Benevolence & 0.645 & $< 10^{-35}$ & 290 \\
PPTS: Interpers.\ Manip. & 0.598 & $< 10^{-29}$ & 290 \\
PPTS: Affective Resp. & 0.594 & $< 10^{-29}$ & 290 \\
SIAS & 0.561 & $< 10^{-25}$ & 290 \\
PPTS: Egocentricity & 0.544 & $< 10^{-24}$ & 290 \\
Trust: Integrity & 0.487 & $< 10^{-18}$ & 290 \\
Trust: Ability & 0.480 & $< 10^{-18}$ & 290 \\
Trust: Propensity & 0.420 & $< 10^{-14}$ & 290 \\
PPTS: Cognitive Resp. & 0.314 & $< 10^{-7}$ & 290 \\
\bottomrule
\end{tabular}
\end{center}
\caption{Beyond-HEXACO subscale recovery ($N = 290$, GPT-4.1 generator, Sonnet scorer). All 9 subscales are recovered at $p < .0033$ (Bonferroni-corrected $\alpha = 0.05/15$).}
\label{tab:beyond_hexaco}
\end{table}

Interpersonal and affective constructs (Trust Benevolence, PPTS Affective Responsiveness, SIAS) recover most strongly, consistent with the finding that personality signal in narratives is carried by content features rather than structural ones (Appendix~\ref{app:content}). PPTS Cognitive Responsiveness recovers most weakly ($r = 0.314$), consistent with this construct measuring interactive mentalising behaviour rather than autobiographical content; however, at $N = 290$ this reaches significance, unlike at $N = 154$ where it did not.

\section{Convergent validity: personality--behaviour correlations}
\label{app:convergent}

\paragraph{Narrative feature $\times$ personality (within-narrative convergent validity):}

To test whether narrative content features reflect the personality profiles that generated them, we correlated 10 features with HEXACO ground truth ($N = 290$, GPT-4.1 narratives, Gemini--Haiku coder average). Table~\ref{tab:convergent_narr} reports the strongest HEXACO predictor per feature; 55 of 204 feature--domain pairs survive Bonferroni correction.

\begin{table}[h]
\begin{center}
\begin{tabular}{llcc}
\toprule
\textbf{Feature} & \textbf{Best HEXACO predictor} & \textbf{$r$} & \textbf{$p$} \\
\midrule
Vulnerability & Emotionality & 0.744 & $< 10^{-50}$ \\
Emotional intensity & Emotionality & 0.576 & $< 10^{-26}$ \\
Warmth & Agreeableness & 0.584 & $< 10^{-27}$ \\
Disclosure depth & Emotionality & 0.585 & $< 10^{-27}$ \\
Emotional complexity & Emotionality & 0.615 & $< 10^{-31}$ \\
Communion & Agreeableness & 0.526 & $< 10^{-21}$ \\
Meaning-making & Honesty-Humility & 0.499 & $< 10^{-18}$ \\
Agency & Extraversion & 0.494 & $< 10^{-18}$ \\
Emotional valence & Extraversion & 0.526 & $< 10^{-21}$ \\
Dominance & Extraversion & 0.312 & $< 10^{-7}$ \\
\bottomrule
\end{tabular}
\end{center}
\caption{Strongest personality predictor per narrative content feature ($N = 290$, GPT-4.1 narratives, Gemini--Haiku coder average).}
\label{tab:convergent_narr}
\end{table}

\paragraph{Cross-context correlations (narrative $\times$ real conversation, same person):}

\begin{table}[h]
\begin{center}
\begin{tabular}{lccc}
\toprule
\textbf{Feature} & \textbf{$r$} & \textbf{$p$} & \textbf{$N$} \\
\midrule
Vulnerability & 0.268 & $< .001$ & 248 \\
Agency & 0.245 & $< .001$ & 248 \\
Emotional valence & 0.218 & $< .001$ & 248 \\
Dominance & 0.210 & .001 & 248 \\
Emotional intensity & 0.195 & .002 & 248 \\
Communion & 0.168 & .008 & 248 \\
Warmth & 0.163 & .010 & 248 \\
Emotional complexity & 0.157 & .013 & 248 \\
Disclosure depth & 0.131 & .039 & 248 \\
Humor & 0.084 & .188 & 248 \\
\bottomrule
\end{tabular}
\end{center}
\caption{Cross-context correlations between content features coded in LLM-generated LSI narratives and real PARSEL conversations for the same individuals ($N = 248$ with $\geq$3 conversations, GPT-4.1 generator, Gemini--Haiku coder average). Nine of ten features are significant (Bonferroni $\alpha = 0.005$, 10 tests).}
\label{tab:convergent_cross}
\end{table}

\paragraph{Multi-generator behavioural replication:}

To confirm that behavioural differentiation is not specific to GPT-4.1, we coded narratives from all three primary generators ($N = 290$ each) using the same coder pair (Gemini + Haiku). Table~\ref{tab:multigenerator} reports the key feature--personality correlations across generators.

\begin{table}[h]
\begin{center}
\small
\begin{tabular}{llcccc}
\toprule
\textbf{Feature} & \textbf{Domain} & \textbf{GPT-4.1} & \textbf{Gem.\ Flash} & \textbf{Grok Fast} & \textbf{Mercury 2} \\
\midrule
Vulnerability & EM & 0.744 & 0.750 & 0.683 & 0.607 \\
Emotional intensity & EM & 0.576 & 0.657 & 0.496 & 0.431 \\
Warmth & AG & 0.584 & 0.585 & 0.557 & 0.468 \\
Communion & AG & 0.526 & 0.520 & 0.515 & 0.441 \\
Agency & EX & 0.494 & 0.519 & 0.411 & 0.391 \\
Meaning-making & HH & 0.499 & 0.477 & 0.492 & 0.317 \\
Creativity/art & OP & 0.455 & 0.432 & 0.466 & 0.402 \\
Emot.\ valence & EX & 0.526 & 0.495 & 0.556 & 0.409 \\
\midrule
\multicolumn{2}{l}{Bonferroni sig.\ pairs} & 55/204 & 51/204 & 59/204 & 41/204 \\
\bottomrule
\end{tabular}
\end{center}
\caption{Key feature--personality correlations across four generators ($N = 290$ each, Gemini--Haiku coder average). Mercury 2 is a diffusion language model; all others are autoregressive. The same features load on the same domains regardless of generator or generation paradigm, with Mercury 2 showing attenuated but directionally consistent signal.}
\label{tab:multigenerator}
\end{table}

The emotional reactivity finding also replicates across all four generators: within-narrative emotional valence variance correlates with Emotionality at $r = 0.303$ (GPT-4.1), $r = 0.312$ (Gemini Flash), $r = 0.291$ (Grok Fast), and $r = 0.249$ (Mercury 2), all $p < 0.001$. The behavioural differentiation is a property of the personality conditioning pipeline, not of any single generator or generation paradigm.

\paragraph{Cross-context by generator:}

\begin{table}[h]
\begin{center}
\small
\begin{tabular}{lcccc}
\toprule
\textbf{Feature} & \textbf{GPT-4.1} & \textbf{Gem.\ Flash} & \textbf{Grok Fast} & \textbf{Mercury 2} \\
\midrule
Vulnerability & .243 & .260 & .230 & .181 \\
Agency & .207 & .115 & .149 & .207 \\
Emot.\ intensity & .161 & .129 & .183 & .215 \\
Emot.\ valence & .190 & .154 & .182 & .138 \\
Dominance & .200 & .190 & .166 & .191 \\
Emot.\ complexity & .137 & .171 & .193 & .123 \\
Warmth & .116 & .101 & .153 & .123 \\
Communion & .116 & .104 & .117 & .153 \\
Disclosure depth & .140 & .062 & .074 & .116 \\
Humor & .082 & .109 & $-$.017 & .039 \\
\midrule
Mean $|r|$ & .159 & .139 & .146 & .149 \\
Sig.\ (of 10) & 7 & 5 & 7 & 6 \\
\bottomrule
\end{tabular}
\end{center}
\caption{Cross-context correlations (narrative vs.\ real conversation) by generator ($N = 248$). Mercury 2 is a diffusion language model. Vulnerability is the strongest bridge feature for all generators. All four generators produce a positive cross-context bridge, confirming the result generalises across generation paradigms.}
\label{tab:cross_context_generators}
\end{table}

\section{Unsupervised narrative topology (BERTopic)}
\label{app:bertopic}

To test whether personality organises narrative structure without supervised feature extraction, we applied BERTopic \citep{Grootendorst2022} to section-level narrative documents. Each of the 24 McAdams sections for all 290 participants was treated as a separate document (6,960 documents total, mean 333 words), embedded with a sentence-transformer model, reduced via UMAP, and clustered with HDBSCAN.

Nineteen topics emerged with only 0.9\% outlier documents, indicating strong topical structure. Correlating topic prevalence per participant with HEXACO ground truth yielded 37 significant personality associations (Bonferroni-corrected). Honesty-Humility emerged as the dominant narrative organising variable: fairness-themed narratives correlated with HH at $r = .478$ (the strongest single topic--personality correlation), and loss narratives (D3) split along the HH/Agreeableness axis, with low-HH/low-A narrators writing about relationship betrayals and high-HH/high-A narrators writing about death and mortality. The E4 (Core Value) prompt split into six distinct value clusters, each with different personality loadings.

The most theoretically informative split occurred on D1 (Life Challenge). Two clusters emerged from the same prompt: one characterised by language of ongoing struggle (``carrying the entire weight,'' ``chronic ever-present anxiety,'' ``a knot in my chest''; positively correlated with Emotionality, $r = .413$) and one characterised by retrospective narration from a position of resolution (``learning to balance my fierce independence,'' ``accepting that sometimes progress requires...''; negatively correlated with Emotionality, $r = -.292$). The embedding space captured narrative \textit{posture}---whether the narrator is still inside the experience or describing it from shore---a distinction invisible to dictionary-based methods.

For comparison, the same pipeline applied to 1,117 speaker-level human conversation documents yielded 13 topics with 47.5\% outliers and a maximum personality correlation of $r = .153$. The contrast reinforces the finding from supervised coding (\S4.3): structured first-person narrative provides sufficient canvas for personality to emerge as topology, while brief conversations do not.

\end{document}